\documentclass[runningheads]{llncs}
\usepackage[T1]{fontenc}
\usepackage{amsmath,bbm}
\usepackage{bm}
\usepackage{booktabs}
\usepackage{graphicx}
\graphicspath{{figures/}}
\DeclareGraphicsExtensions{.pdf,.png}

\usepackage[symbol]{footmisc}

\newcommand*\samethanks[1][\value{footnote}]{\footnotemark[#1]}

\begin{document}
\newcommand{\MNAME}{OPIC}
\title{Zero-shot Learning of Individualized Task Contrast Prediction from Resting-state Functional Connectomes}
\titlerunning{Zero-shot Learning of Individualized Task Contrast Prediction from rsFC}
\author{Minh Nguyen\inst{1}\thanks{indicates equal contribution} \and
Gia H. Ngo\inst{1}\samethanks~\and
Mert R. Sabuncu\inst{1,2}}
\authorrunning{Nguyen \& Ngo et al.}
\institute{School of Electrical \& Computer Engineering, Cornell University, USA \and
Radiology, Weill Cornell Medicine, USA}
\maketitle
\begin{abstract}
Given sufficient pairs of resting-state and task-evoked fMRI scans from subjects, it is possible to train ML models to predict subject-specific task-evoked activity using resting-state functional MRI (rsfMRI) scans.
However, while rsfMRI scans are relatively easy to collect, obtaining sufficient task fMRI scans is much harder as it involves more complex experimental designs and procedures.
Thus, the reliance on scarce paired data limits the application of current techniques to only tasks seen during training.
We show that this reliance can be reduced by leveraging group-average contrasts, enabling zero-shot predictions for novel tasks.
Our approach, named \MNAME~(short for Omni-Task Prediction of Individual Contrasts), takes as input a subject's rsfMRI-derived connectome and a group-average contrast, to produce a prediction of the subject-specific contrast.
Similar to zero-shot learning in large language models using special inputs to obtain answers for novel natural language processing tasks, inputting group-average contrasts guides the \MNAME~model to generalize to novel tasks unseen in training.
Experimental results show that \MNAME's predictions for novel tasks are not only better than simple group-averages, but are also competitive with a state-of-the-art model's in-domain predictions that was trained using in-domain tasks' data.
\keywords{out-of-domain \and zero-shot \and functional connectivity \and task-induced fingerprint}
\end{abstract}

\section{Introduction}
\begin{figure}
\centering
\includegraphics[width=.79\textwidth]{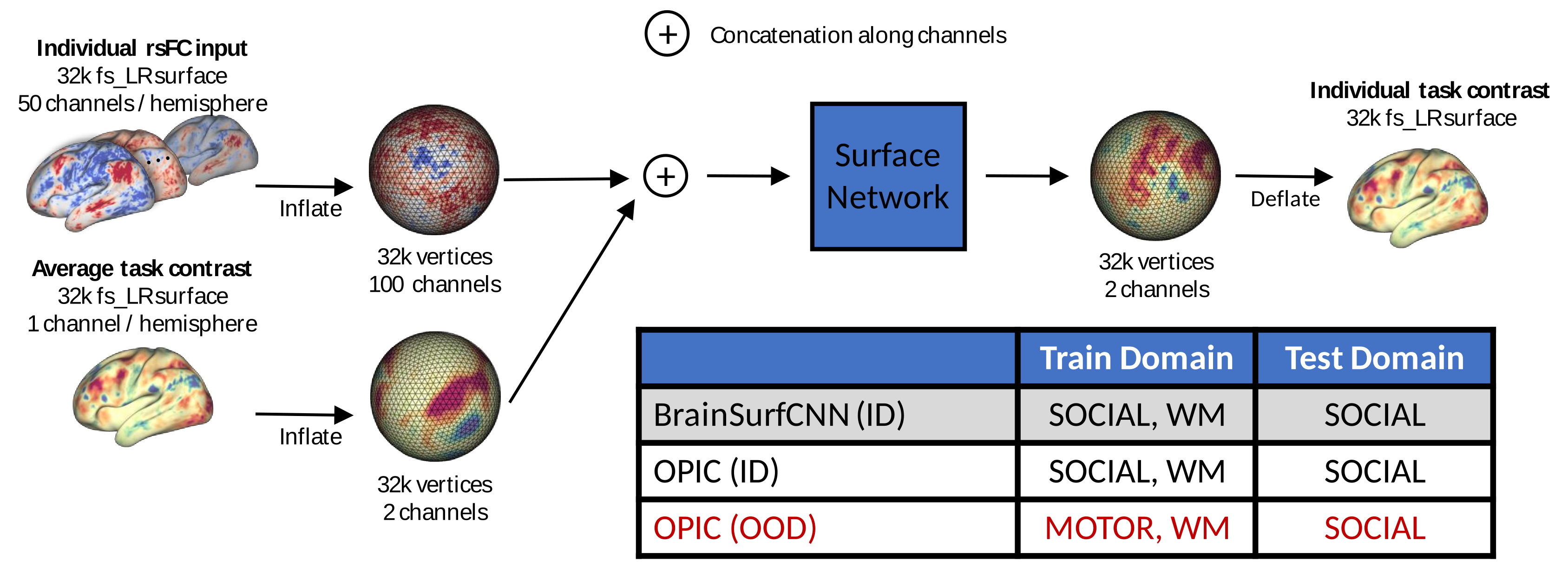}
\caption{\MNAME~predicts individualized task-evoked contrast using group-average contrast and subject's rsFC.
\MNAME~is based on BrainSurfCNN which is a surface-based CNN with spherical convolutional kernels.
However, \MNAME~can predict both ID (e.g.~trained and tested on SOCIAL contrasts) and OOD individualized contrasts (e.g.~trained on WM and MOTOR but tested on SOCIAL contrasts as shown in red) while BrainSurfCNN is not capable of OOD prediction.
ID: in-domain, OOD: out-of-domain.}\label{fig:model}
\end{figure}
Functional connectomes derived from resting-state functional MRI (rsfMRI) scans contain subject-specific characteristics that could be used as ``fingerprints'' of individual cognitive functions~\cite{biswal2010toward,kelly2012characterizing}.
Such fingerprints have many uses~\cite{khosla2019machine}, for example predicting individual developmental trajectories~\cite{dosenbach2010prediction}, behavioral traits~\cite{finn2015functional,pang2022resting}, or individualized task-evoked activities~\cite{tavor2016task,cole2016activity,zheng2022accurate,ngo2020connectomic}.
The last one is useful in pre-surgical planning or studying of neurological disorders~\cite{dimou2013systematic,castellano2017functional,salama2018diffusion,bernstein2022prediction} and has been tackled using various methods, e.g.~GLM~\cite{tavor2016task}, ensembles~\cite{zheng2022accurate}, or neural networks~\cite{ngo2020connectomic,ngo2022predicting}.
For each task-evoked activity, existing methods need paired task fMRI-rsfMRI scans as training data.
Thus, they cannot predict for activities unseen in training data (out-of-domain generalization).
Besides, paired training data is hard to collect because it is hard to reliably collect high quality task fMRI scans~\cite{elliott2020test} (due to poor task performance and measurement noise).
The lack of out-of-domain generalization and the scarcity of paired training data limit the potential utility of these methods.
Meta-learning or few-shot learning~\cite{andrychowicz2016learning,ravi2016optimization,finn2017model} can improve generalization given some paired data from out-of-domain activities.
However, when no additional paired data is available, generalizing to unseen task-evoked activities requires zero-shot learning~\cite{larochelle2008zero,yu2010attribute,rohrbach2011evaluating,lampert2013attribute}.

We propose \MNAME~(short for Omni-Task Prediction of Individual Contrasts), an approach that relaxes the need for paired training data and enables zero-shot predictions for novel tasks.
\MNAME~takes as input a subject's rsfMRI-derived connectome and a group-average contrast, and predicts the subject-specific task-evoked activity (contrast).
The group-average controls the task identity while the rsfMRI input determines subject-specific patterns in the contrast.
\MNAME~can make individualized predictions from rsfMRI connectomes for both in-domain and out-of-domain contrasts.
Although zero-shot learning using additional model's input in a multitask setting~\cite{caruana1997multitask} has been widely used in natural language processing~\cite{brown2020language,raffel2020exploring}, as far as we know, we are the first to examine its utility in our application domain.
Our experiments with the HCP dataset~\cite{glasser2013minimal} show that the proposed \MNAME's out-of-domain prediction is better than a simple group-average and competitive with a state-of-the-art (SOTA) model's in-domain prediction that requires paired training data.
\MNAME's out-of-domain prediction also matches the SOTA model's in-domain prediction in subject identification, an important requisite for creating individualized task-evoked fingerprints.

\section{Methods}
Figure~\ref{fig:model} shows the \MNAME~model.
Its input and output are multi-channel icosahedral fs\_LR meshes with 32,492 vertices per brain hemisphere~\cite{van2012parcellations}.
fs\_LR is an atlas based on imaging data (anatomical/structural, resting-state, myelination) from a large pool of subjects to achieve better inter-subject correspondence~\cite{van2012parcellations}.
The output is a subject-specific task-evoked contrast.
\MNAME's surface network is similar to the publicly-available BrainSurfCNN~\cite{ngo2020connectomic}, a U-Net-based~\cite{ronneberger2015u,milletari2016v} surface-based CNN with spherical convolutional kernels~\cite{chiyu2019spherical} (see~\cite{ngo2020connectomic} for the network specifics).
However, \MNAME~differs from BrainSurfCNN in two key ways.

First, while BrainSurfCNN designates 2 specific output channels for each task, \MNAME~uses the same 2 output channels for all tasks (parameter-sharing).
Although BrainSurfCNN can predict for multiple tasks, it cannot predict for a new task because each task uses different output channels and the number of output channels is fixed.
In contrast, through parameter-sharing, \MNAME~can use the same output channels to predict for previously unseen (out-of-domain) tasks.

Second, in addition to the rsfMRI derived functional connectome (rsFC) input, \MNAME~also has a group-average contrast input.
The rsFC consists of functional connectivity features, each of which is the Pearson's correlation between the timeseries of a vertex on the icosahedral mesh and the averaged timeseries of an ROI.
In our experiments, the group-average contrast is the average of all the task contrast maps across (training) subjects for a specific task.
Since the group-average provides a rough estimate of subject-specific task-evoked contrast, this setup encourages \MNAME~to learn to map differences in rsFC to individual differences in task-evoked contrasts.
Thus, for an out-of-domain activity, \MNAME~can refine the group-average input using information from rsFC to produce the individualized task-evoked contrast.
Since this is possible without having to fine-tune using new paired data, this setup reduces the burden of collecting paired data.

\section{Experimental setup}
\subsection{Data}\label{ssec:data}
Our experiments used 3-Tesla resting-state fMRI (rsfMRI) and task fMRI (tfMRI) data of subjects from the Human Connectome Project (HCP)~\cite{glasser2013minimal}.
Data acquisition and pre-processing details are described in~\cite{glasser2013minimal,smith2013resting,barch2013function}.
Each HCP subject has up to four 15-minute runs of rsfMRI data and tfMRI data from 86 tasks belonging to 7 task groups.
The 7 task groups are EMOTION, GAMBLING, LANGUAGE, MOTOR, RELATIONAL, SOCIAL, and WM (working memory)~\cite{barch2013function}.
For each contrast, there is one corresponding group-average contrast.
Following~\cite{tavor2016task,ngo2020connectomic}, we excluded redundant negative tfMRI contrasts (47 tasks remained).
We also excluded subjects with fewer than 4 rsfMRI runs.
Of the remaining subjects, 39 subjects are held out for testing, 774 subjects are used for training and 19 are used for validation.
All test subjects have second visits (retest data).
No family members are split across test and training or validation sets.
The 50-component ICA timeseries data included in HCP are used to compute subjects' functional connectomes.
Each element in a subject's connectome corresponds to the Pearson's correlation between the rsfMRI timeseries at a vertex and the timeseries of an ICA-component.
The group-average map is scaled so that its absolute maximum value is 1, matching the magnitude of the functional connectomes.

\begin{table}[ht]
\caption{Comparing \MNAME's in-domain (ID) and out-of-domain (OOD) performance against the baselines' in term of average AUC.
\MNAME~is significantly better than linear regression (Lin. Regr.) and group-average (Grp. Avg.) prediction.
\MNAME's in-domain performance is on-par with BrainSurfCNN (BSC) for all contrasts.
While it is an unfair comparison for \MNAME, \MNAME's out-of-domain performance is on-par with BrainSurfCNN's in-domain performance in some contrasts (e.g. ``GAMBLING REWARD'' contrast).
\MNAME~is also on-par with the retest session (Retest).\\
*: ID \MNAME~beats baseline, $\dagger$: OOD \MNAME~beats baseline (paired 2-tail t-test at $p{<}1e{-4}$)
}\label{tbl:id_ood_result}
\begin{tabular}{lllllll}
\toprule
           Contrast & \multicolumn{2}{c}{\MNAME} &  \quad BSC  & Grp. Avg. & Lin. Regr. & Retest \\
                    & \quad  ID  & \quad  OOD &   & & & \\
\midrule                                
    GAMBLING PUNISH & \quad0.287 & \quad0.284 & \quad0.287  &  0.272*$\dagger$ &  0.259*$\dagger$ &  0.274* \\
    GAMBLING REWARD & \quad0.296 & \quad0.293 & \quad0.295  &  0.279*$\dagger$ &  0.266*$\dagger$ &  0.281* \\
LANGUAGE MATH-STORY & \quad0.287 & \quad0.284 & \quad0.288  &  0.272*$\dagger$ &  0.269*$\dagger$ &  0.286  \\
          MOTOR CUE & \quad0.257 & \quad0.253 & \quad0.255  &  0.240*$\dagger$ &  0.218*$\dagger$ &  0.243* \\
      MOTOR CUE-AVG & \quad0.254 & \quad0.250 & \quad0.252  &  0.241*$\dagger$ &  0.222*$\dagger$ &  0.248  \\
   RELATIONAL MATCH & \quad0.311 & \quad0.307 & \quad0.310  &  0.294*$\dagger$ &  0.279*$\dagger$ &  0.305  \\
     RELATIONAL REL & \quad0.317 & \quad0.312 & \quad0.316  &  0.299*$\dagger$ &  0.284*$\dagger$ &  0.310  \\
      SOCIAL RANDOM & \quad0.302 & \quad0.300 & \quad0.299* &  0.289*$\dagger$ &  0.259*$\dagger$ &  0.302  \\
         SOCIAL TOM & \quad0.308 & \quad0.306 & \quad0.308  &  0.292*$\dagger$ &  0.266*$\dagger$ &  0.312  \\
             WM 0BK & \quad0.296 & \quad0.291 & \quad0.295  &  0.280*$\dagger$ &  0.266*$\dagger$ &  0.281* \\
             WM 2BK & \quad0.305 & \quad0.299 & \quad0.304  &  0.285*$\dagger$ &  0.271*$\dagger$ &  0.297  \\
            WM FACE & \quad0.282 & \quad0.277 & \quad0.281  &  0.266*$\dagger$ &  0.249*$\dagger$ &  0.268* \\
\bottomrule
\end{tabular}
\end{table}
\subsection{\MNAME's training}
To test both in-domain and out-of-domain generalization, \MNAME~was trained in a leave-one-group-out manner.
Specifically, we trained \MNAME~seven times and each time the model was trained from scratch with one task group excluded from the training data.
For out-of-domain evaluation, the model was tested on the excluded task group.
For in-domain evaluation, the model was tested on the task groups seen during training, but on independent test subjects.
Since there are six different predictions for a subject's task contrast, we averaged the predictions.
The \MNAME~model was trained to minimize the $l^2$ loss using the Adam optimizer~\cite{kingma2015adam}.
It was trained for 20 epochs with batch size 10 and learning rate 3E-3.
Training the \MNAME~model took 1 day using a NVIDIA Titan Xp GPU.
The iteration with the lowest validation error was chosen for evaluation.

\subsection{Baselines}
\begin{figure}[ht]
\centering
\includegraphics[width=.90\textwidth]{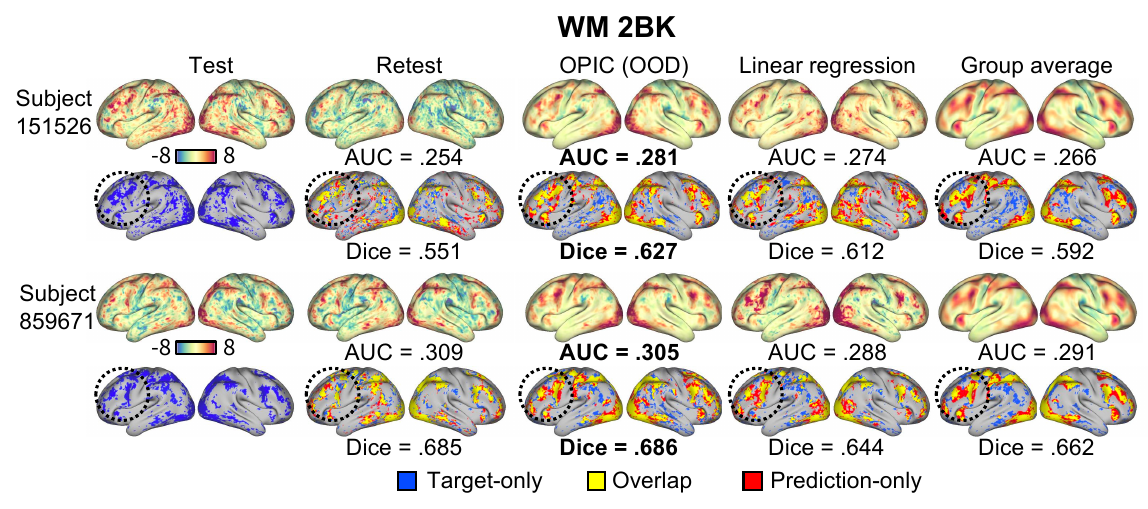}
\caption{Surface visualization for a representative task contrast (``WM 2BK'') of 2 subjects.
\MNAME~prediction is out-of-domain (Section~\ref{ref:subsec_ood}) while linear regression was trained with the target contrast on the training subjects (in-domain).
For each subject, the top row shows the unthresholded prediction or reference, and the bottom row show the overlap between the prediction or reference and the target.
The right-most column shows the group-average contrasts for comparison.}\label{fig:wm_example}
\end{figure}
We compared the \MNAME~model against 4 baselines: the state-of-the-art BrainSurfCNN model, a linear regression model, the group-average, and the retest scan.
The BrainSurfCNN model's training follows~\cite{ngo2020connectomic} whereby the model was first trained with $l^2$ loss and then finetuned using the R-C loss.
The R-C loss~\cite{ngo2020connectomic} minimizes the error reconstructing an individual task contrast from their functional connectome and maximizes the difference with other subjects' contrasts.
The linear regression is set up similar to~\cite{tavor2016task,ngo2020connectomic} whereby there is one regressor for each parcel (parcellation was derived from group-level ICA; Section~\ref{ssec:data}).  
For each parcel and task, subject-specific regressors were fitted and the weights of fitted regressors of subjects from the training and validation set were averaged to create an average regressor used for prediction.
The group-average map is a naive baseline that ignores inter-subject variability.
This baseline would be inadequate for tasks with high inter-subject variability and thus is considered a lower-bound on performance.
On the other hand, outputting the retest (repeat) scan as the prediction would be an effective upper-bound on performance since
there should be high consistency between two scans of the same subject~\cite{ngo2020connectomic}.

\subsection{Metrics}
We computed Dice scores~\cite{dice1945measures} at different thresholds, where the overlap between the top $X$-percent (where $X$ is varied between 5 and 50) of vertices in the predicted and target (tfMRI-derived, observed) contrast maps was quantified. 
This strategy allows us to assess the quality of the prediction at different levels of detail~\cite{ngo2020connectomic}.
The area under the Dice curve (AUC) across all thresholds is used to measure how similar the models' predictions are to the target contrast.

In quantifying \MNAME's prediction accuracy, we consider three different scenarios.
First, we examine the relatively easy problem of in-domain prediction, where the test-time task was seen in training.
Second, we consider the more difficult problem of out-of-domain prediction, where the performance is assessed for tasks that were not part of the groups seen in training.
Finally, in the third scenario, we consider test-time tasks that are part of a group that was previously seen but the specific task was not included in the training.

\begin{figure}[ht]
\centering
\includegraphics[width=.90\textwidth]{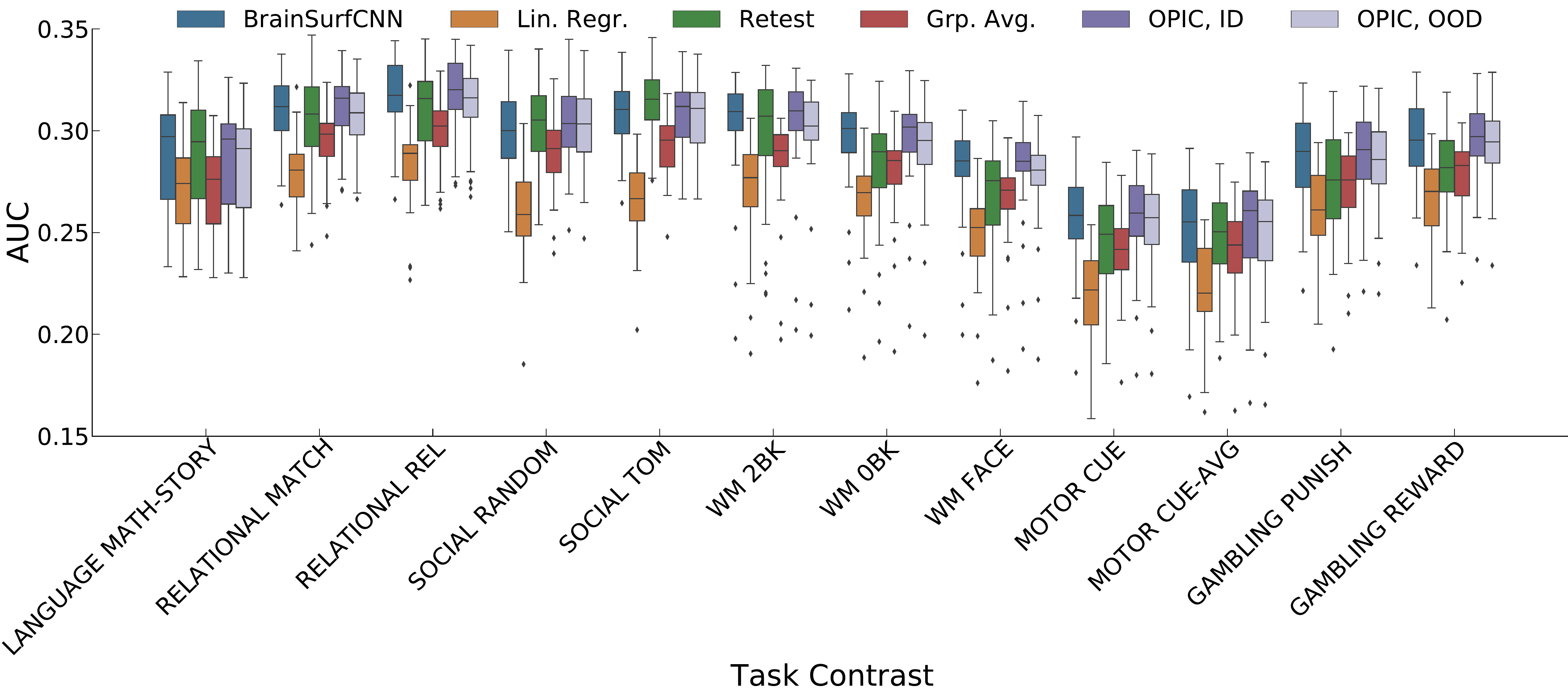}
\caption{Average AUC for reliably predictable tasks.
\MNAME's in-domain prediction (\MNAME, ID) is on-par with BrainSurfCNN's in-domain prediction.
Both \MNAME's in-domain prediction (\MNAME, ID) and out-of-domain prediction (\MNAME, OOD) are consistently better than in-domain prediction from linear regression (Lin. Regr.) and group-average (Grp. Avg.) prediction.
\MNAME's performance is on-par with the retest session (Retest). 
See Table~\ref{tbl:id_ood_result} for more details.
}\label{fig:id_ood_result}
\end{figure}
In addition to AUC, we used ``subject identification accuracy'' that measures how specific the predicted contrasts are to individual subjects~\cite{tavor2016task,ngo2020connectomic}.
In this analysis, for each subject, the AUC between the subject's predicted and observed contrasts of all subjects is first computed.
This results in a square identification matrix with a dimension equal to the number of test subjects.
Subject identification accuracy is in turn computed as the fraction of subjects whereby the subject's prediction accurately identifies the subject, i.e., accuracy is the average row-wise sorted rank order of the diagonal elements.
Following~\cite{tavor2016task,ngo2020connectomic}, the identification matrices were normalized before computing accuracy.

\section{Results}
\subsection{In-domain prediction quality}
Figure~\ref{fig:id_ood_result} and Table~\ref{tbl:id_ood_result} show the comparison between \MNAME's in-domain performance against the baselines for reliably predictable tasks.
Following~\cite{ngo2020connectomic}, a predictable task is defined as one in which a subject's tfMRI test scan is more similar to their retest scan (on average, measured by AUC), compared to the group-average map.
The results for \textit{all} tasks are included in the Supplementary Material.
The results show that \MNAME's performance matches BrainSurfCNN, and is on par with test-retest reliability.
\MNAME~consistently outperforms the group-average baseline and the linear regression baseline.

\begin{figure}[ht]
\centering
\includegraphics[width=.90\textwidth]{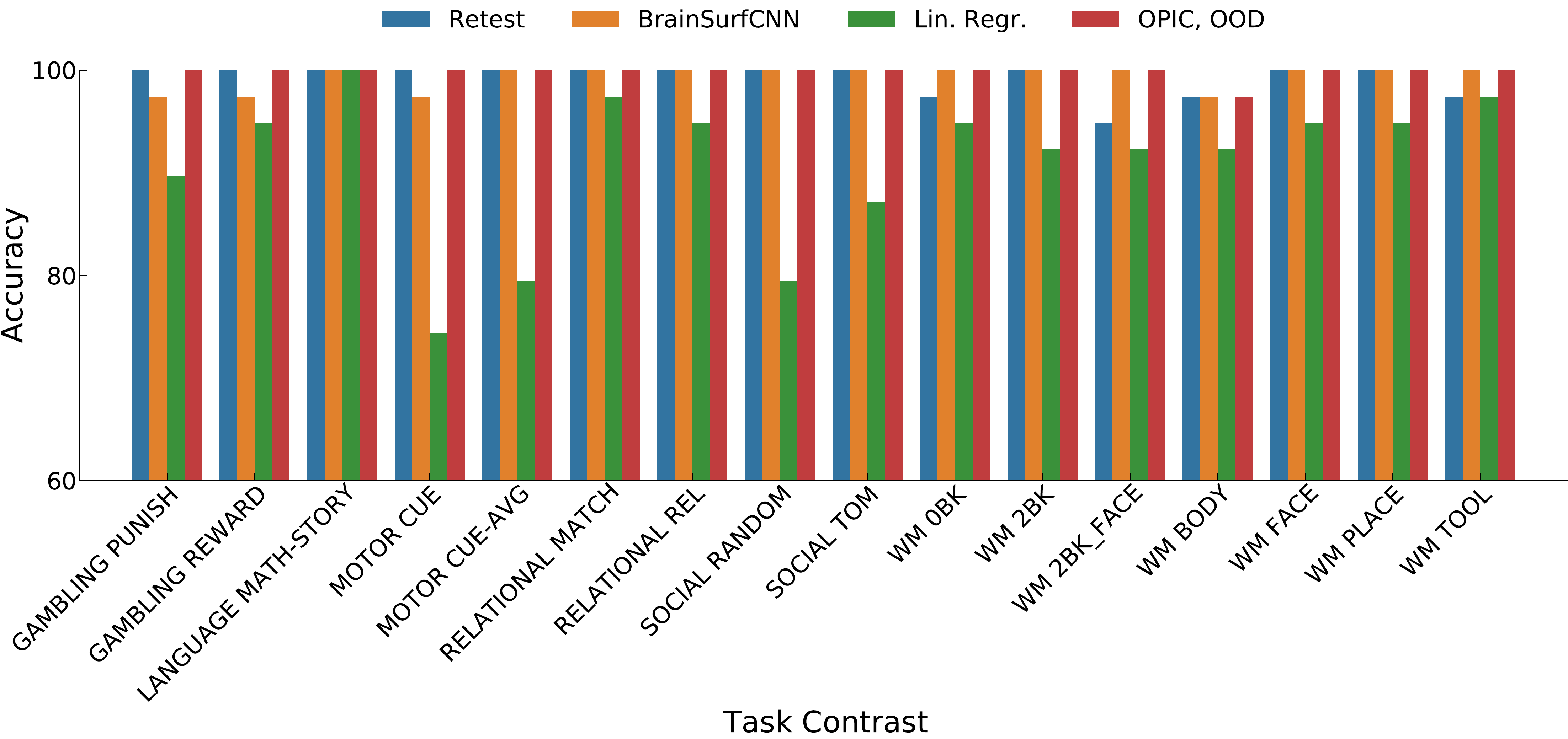}
\caption{Subject identification accuracy for reliably predictable tasks.
\MNAME~(OOD) is on-par with BrainSurfCNN (ID) even on tasks \MNAME~was not trained to predict.}\label{fig:identifiability}
\end{figure}
\subsection{Out-of-domain prediction quality}\label{ref:subsec_ood}
Figure~\ref{fig:id_ood_result} and Table~\ref{tbl:id_ood_result} also show the comparison between \MNAME's out-of-domain (OOD) prediction and the baselines' performance.
While \MNAME~was tested on tasks from a group unseen in training, the linear regression baseline and the BrainSurfCNN baseline can only make predictions for tasks seen in training.

Figure~\ref{fig:wm_example} shows the prediction of a representative task contrast, ``WM 2BK'', (which has median average target-retest Dice AU), for 2 test subjects with the 10th and 90th percentile target-group average Dice AUCs.
Note that \MNAME~prediction is from a model trained with the OOD setup, i.e.\ it was not trained on any ``WM'' task in the HCP dataset, while linear regression model was estimated from training subjects on the target ``WM 2BK'' task.
Despite the handicap, \MNAME~outperforms the linear regression baseline and group-average reference in Dice score for the top 25\% most activated vertices, and the overall Dice AUC score of the unthresholded maps.
Focusing on the prefrontal cortex (circled), although the group-average contrast coarsely overlaps with the subjects' activation, \MNAME~prediction significantly matches the subject-specific minute variation.
This suggests that \MNAME~does not merely copy the group-average contrast of the training subjects, but indeed makes subject-specific prediction.

The quantitative results also show that \MNAME~consistently outperforms the group-average baseline.
Thus, \MNAME~is able to combine the group-average contrast input with the subject-specific rsFC input to accurately predict a subject-specific contrast map.
In addition, \MNAME~is consistently better than the linear regression's in-domain prediction and is on-par with the retest AUC.
These results demonstrate \MNAME's ability to generalize well to new tasks from previously unseen groups.
Comparing \MNAME's OOD prediction with BrainSurfCNN in-domain prediction is not exactly fair for \MNAME~because \MNAME~only knows about a prediction task (via the group-average contrast) during test time while BrainSurfCNN was trained to perform the task.
Yet, we observe that \MNAME~manages to reach BrainSurfCNN's level of performance in some contrasts such as ``GAMBLING REWARD'' or ``SOCIAL TOM'' (Table~\ref{tbl:id_ood_result}).

\begin{table}[ht]
\caption{\MNAME~generalizes well to OOD contrasts, almost matching that of in-domain predictions. AUC values are listed. P-values are for paired 2-tail t-test comparison.
}\label{tbl:ablate}
\begin{tabular}{lccc}
\toprule
    Task & \quad In-domain \quad & \quad OOD, seen group \quad & \quad OOD, new group \quad \\
\midrule
WM PLACE-AVG & \quad0.209 (p=7e-2)\quad & 0.208 & \quad 0.206 (p=3e-3) \quad \\
 WM TOOL-AVG & \quad0.178 (p=3e-3)\quad & 0.177 & \quad 0.177 (p=8e-1) \quad \\
\bottomrule
\end{tabular}
\end{table}
Figure~\ref{fig:identifiability} shows the subject identification accuracy of predictions for reliably predictable task contrasts.
The \MNAME~model prediction is out-of-domain (unseen tasks) while the BrainSurfCNN's and linear regression's prediction is in-domain (tasks seen during training).
\MNAME~matches BrainSurfCNN in subject identifiability even on contrasts \MNAME~was not trained to predict.
Thus, \MNAME~seems to produce accurate predictions of individualized task contrast maps.

\subsection{New task contrast from a seen task group}
In practice, a model is more likely to encounter an unseen task from a seen group than a completely new group.
We compared \MNAME's performance in this scenario against its in-domain prediction results.
Specifically, we trained an \MNAME~model on all the available contrasts except two, ``WM PLACE-AVG'' and ``WM TOOL-AVG''.
Table~\ref{tbl:ablate} shows that \MNAME's performance on a new task from a seen group is better than a new task from an unseen group, but slightly worse than the scenario where the task was seen during training (in-domain).

\section{Conclusion}
Accurate prediction of individual differences in task-evoked brain activities~\cite{tavor2016task,cole2016activity,zheng2022accurate,ngo2022predicting} using cognitive fingerprints extracted from rsfMRI~\cite{finn2015functional,finn2021beyond} has several potential clinical applications~\cite{dimou2013systematic,castellano2017functional,salama2018diffusion,bernstein2022prediction}.
Alas, prior approaches needs paired rsfMRI-task-fMRI scans for model fitting but paired data are scarce because of measurement noise and subjects' poor task performance.
Predicting individualized task-evoked activities for novel (out-of-domain) tasks that have no paired data would require zero-shot generalization.
In this paper, we proposed an approach to make individualized prediction for both in-domain and out-of-domain tasks by additionally leveraging group-average contrasts.
Our \MNAME~model not only is on-par with competitive baselines in in-domain setting, but also can predict out-of-domain activities, something that prior approaches cannot do.

Although having presented a wide range of experimental settings, there are many unexplored areas left for future work.
First, this work shows out-of-domain tasks in the same dataset.
Can \MNAME~predict out-of-domain tasks across different datasets or different populations (e.g.\ different age groups, patients vs.\ controls)?
Verifying \MNAME's reliability when applying to populations with some cognitive or mental health issues would be important before clinical adoption.
Second, can \MNAME~predict using imperfect group-average contrasts, for example group-average contrasts from meta-analyses~\cite{poldrack2016brain,ngo2021text,ngo2022transformer}?
Third, we did not use the contrastive R-C loss in training, as this loss function was originally designed for a model that predicts multiple contrasts from the same input concurrently.
However, it would be interesting to study if some form of contrastive objective further improves predictive performance while retaining inter-subject variability.

\bibliographystyle{splncs04}
\bibliography{main}

\begin{thebibliography}{10}
\providecommand{\url}[1]{\texttt{#1}}
\providecommand{\urlprefix}{URL }
\providecommand{\doi}[1]{https://doi.org/#1}

\bibitem{andrychowicz2016learning}
Andrychowicz, M., Denil, M., Gomez, S., Hoffman, M., Pfau, D., Schaul, T.,
  Shillingford, B., De~Freitas, N.: Learning to learn by gradient descent by
  gradient descent. NIPS  \textbf{29} (2016)

\bibitem{barch2013function}
Barch, D., Burgess, G., Harms, M., Petersen, S., Schlaggar, B., Corbetta, M.,
  Glasser, M., Curtiss, S., Dixit, S., Feldt, C., et~al.: {Function in the
  human connectome: task-fMRI and individual differences in behavior}.
  NeuroImage  \textbf{80} (2013)

\bibitem{bernstein2022prediction}
Bernstein-Eliav, M., Tavor, I.: The prediction of brain activity from
  connectivity: Advances and applications. The Neuroscientist  (2022)

\bibitem{biswal2010toward}
Biswal, B., Mennes, M., Zuo, X.N., Gohel, S., Kelly, C., Smith, S., Beckmann,
  C., Adelstein, J., Buckner, R., Colcombe, S., et~al.: Toward discovery
  science of human brain function. PNAS  \textbf{107}(10) (2010)

\bibitem{brown2020language}
Brown, T., Mann, B., Ryder, N., Subbiah, M., Kaplan, J., Dhariwal, P.,
  Neelakantan, A., et~al.: Language models are few-shot learners. In: NeurIPS
  (2020)

\bibitem{caruana1997multitask}
Caruana, R.: Multitask learning. Machine learning  \textbf{28}(1) (1997)

\bibitem{castellano2017functional}
Castellano, A., Cirillo, S., Bello, L., Riva, M., Falini, A.: Functional mri
  for surgery of gliomas. Current treatment options in neurology
  \textbf{19}(10) (2017)

\bibitem{chiyu2019spherical}
Chiyu, M., Huang, J., Kashinath, K., Prabhat, M., Niessner, M.: Spherical cnns
  on unstructured grids. In: ICLR (2019)

\bibitem{cole2016activity}
Cole, M., Ito, T., Bassett, D., Schultz, D.: Activity flow over resting-state
  networks shapes cognitive task activations. Nature Neuroscience
  \textbf{19}(12) (2016)

\bibitem{dice1945measures}
Dice, L.: Measures of the amount of ecologic association between species.
  Ecology  \textbf{26}(3) (1945)

\bibitem{dimou2013systematic}
Dimou, S., Battisti, R., Hermens, D.F., Lagopoulos, J.: A systematic review of
  functional mri and dti modalities used in presurgical planning of brain
  tumour resection. Neurosurgical Review  \textbf{36}(2) (2013)

\bibitem{dosenbach2010prediction}
Dosenbach, N., Nardos, B., Cohen, A., Fair, D., Power, J., Church, J., et~al.:
  {Prediction of individual brain maturity using fMRI}. Science
  \textbf{329}(5997) (2010)

\bibitem{elliott2020test}
Elliott, M., Knodt, A., Ireland, D., Morris, M., Poulton, R., Ramrakha, S.,
  Sison, M., Moffitt, T., Caspi, A., Hariri, A.: What is the test-retest
  reliability of common task-functional mri measures? new empirical evidence
  and a meta-analysis. Psychological science  \textbf{31}(7) (2020)

\bibitem{finn2017model}
Finn, C., Abbeel, P., Levine, S.: Model-agnostic meta-learning for fast
  adaptation of deep networks. In: ICML. PMLR (2017)

\bibitem{finn2021beyond}
Finn, E., Rosenberg, M.: Beyond fingerprinting: Choosing predictive connectomes
  over reliable connectomes. NeuroImage  \textbf{239} (2021)

\bibitem{finn2015functional}
Finn, E., Shen, X., Scheinost, D., Rosenberg, M., Huang, J., Chun, M.,
  Papademetris, X., Constable, T.: Functional connectome fingerprinting:
  identifying individuals using patterns of brain connectivity. Nature
  Neuroscience  \textbf{18}(11) (2015)

\bibitem{glasser2013minimal}
Glasser, M., Sotiropoulos, S., Wilson, A., Coalson, T., Fischl, B., Andersson,
  J., et~al.: The minimal preprocessing pipelines for the human connectome
  project. NeuroImage  \textbf{80} (2013)

\bibitem{kelly2012characterizing}
Kelly, C., Biswal, B., Craddock, C., Castellanos, X., Milham, M.:
  Characterizing variation in the functional connectome: promise and pitfalls.
  {Trends in Cognitive Sciences}  \textbf{16}(3) (2012)

\bibitem{khosla2019machine}
Khosla, M., Jamison, K., Ngo, G., Kuceyeski, A., Sabuncu, M.: {Machine learning
  in resting-state fMRI analysis}. Magnetic Resonance Imaging  (2019)

\bibitem{kingma2015adam}
Kingma, D., Ba, J.: Adam: {A} {Method} for {Stochastic} {Optimization}. In:
  ICLR (2014)

\bibitem{lampert2013attribute}
Lampert, C., Nickisch, H., Harmeling, S.: Attribute-based classification for
  zero-shot visual object categorization. IEEE TPAMI  \textbf{36}(3) (2013)

\bibitem{larochelle2008zero}
Larochelle, H., Erhan, D., Bengio, Y.: Zero-data learning of new tasks. In:
  AAAI. vol.~1 (2008)

\bibitem{milletari2016v}
Milletari, F., Navab, N., Ahmadi, S.A.: V-net: Fully convolutional neural
  networks for volumetric medical image segmentation. In: International
  Conference on 3D Vision. IEEE (2016)

\bibitem{ngo2020connectomic}
Ngo, G., Khosla, M., Jamison, K., Kuceyeski, A., Sabuncu, M.: From connectomic
  to task-evoked fingerprints: Individualized prediction of task contrasts from
  resting-state functional connectivity. In: MICCAI (2020)

\bibitem{ngo2022predicting}
Ngo, G., Khosla, M., Jamison, K., Kuceyeski, A., Sabuncu, M.: Predicting
  individual task contrasts from resting-state functional connectivity using a
  surface-based convolutional network. NeuroImage  \textbf{248} (2022)

\bibitem{ngo2021text}
Ngo, G., Nguyen, M., Chen, N., Sabuncu, M.: Text2brain: Synthesis of brain
  activation maps from free-form text query. In: MICCAI (2021)

\bibitem{ngo2022transformer}
Ngo, G., Nguyen, M., Chen, N., Sabuncu, M.: A transformer-based neural language
  model that synthesizes brain activation maps from free-form text queries.
  Medical Image Analysis  (2022)

\bibitem{pang2022resting}
Pang, L., Li, H., Liu, Q., Luo, Y.J., Mobbs, D., Wu, H.: Resting-state
  functional connectivity of social brain regions predicts motivated
  dishonesty. NeuroImage  \textbf{256} (2022)

\bibitem{poldrack2016brain}
Poldrack, R., Yarkoni, T.: From brain maps to cognitive ontologies: informatics
  and the search for mental structure. Annual review of psychology  \textbf{67}
  (2016)

\bibitem{raffel2020exploring}
Raffel, C., Shazeer, N., Roberts, A., Lee, K., Narang, S., Matena, M., Zhou,
  Y., Li, W., Liu, P.: Exploring the limits of transfer learning with a unified
  text-to-text transformer. JMLR  \textbf{21} (2020)

\bibitem{ravi2016optimization}
Ravi, S., Larochelle, H.: Optimization as a model for few-shot learning. In:
  ICLR (2017)

\bibitem{rohrbach2011evaluating}
Rohrbach, M., Stark, M., Schiele, B.: Evaluating knowledge transfer and
  zero-shot learning in a large-scale setting. In: CVPR. IEEE (2011)

\bibitem{ronneberger2015u}
Ronneberger, O., Fischer, P., Brox, T.: U-net: Convolutional networks for
  biomedical image segmentation. In: MICCAI (2015)

\bibitem{salama2018diffusion}
Salama, G., Heier, L., Patel, P., Ramakrishna, R., Magge, R., Tsiouris, A.:
  Diffusion weighted/tensor imaging, functional mri and perfusion weighted
  imaging in glioblastoma—foundations and future. Frontiers in neurology
  \textbf{8} (2018)

\bibitem{smith2013resting}
Smith, S., Beckmann, C., Andersson, J., Auerbach, E., Bijsterbosch, J., Douaud,
  G., Duff, E., Feinberg, D., Griffanti, L., Harms, M., et~al.: {Resting-state
  fMRI in the Human Connectome Project}. NeuroImage  \textbf{80} (2013)

\bibitem{tavor2016task}
Tavor, I., Jones, P., Mars, R., Smith, S., Behrens, T., Jbabdi, S.: Task-free
  mri predicts individual differences in brain activity during task
  performance. Science  \textbf{352}(6282) (2016)

\bibitem{van2012parcellations}
Van~Essen, D., Glasser, M., Dierker, D., Harwell, J., Coalson, T.:
  Parcellations and hemispheric asymmetries of human cerebral cortex analyzed
  on surface-based atlases. Cerebral Cortex  \textbf{22}(10) (2012)

\bibitem{yu2010attribute}
Yu, X., Aloimonos, Y.: Attribute-based transfer learning for object
  categorization with zero/one training example. In: ECCV (2010)

\bibitem{zheng2022accurate}
Zheng, Y.Q., Farahibozorg, S.R., Gong, W., Rafipoor, H., Jbabdi, S., Smith, S.:
  Accurate predictions of individual differences in task-evoked brain activity
  from resting-state fmri using a sparse ensemble learner. Neuroimage
  \textbf{259} (2022)

\end{thebibliography}

\appendix
\newpage
\pagenumbering{arabic}
\setcounter{page}{1}
\renewcommand{\thefigure}{S\arabic{figure}}
\setcounter{figure}{0}
\begin{center}
\large\textbf{Supplementary Materials}
\end{center}

\begin{figure}
\centering
\includegraphics[width=.90\textwidth]{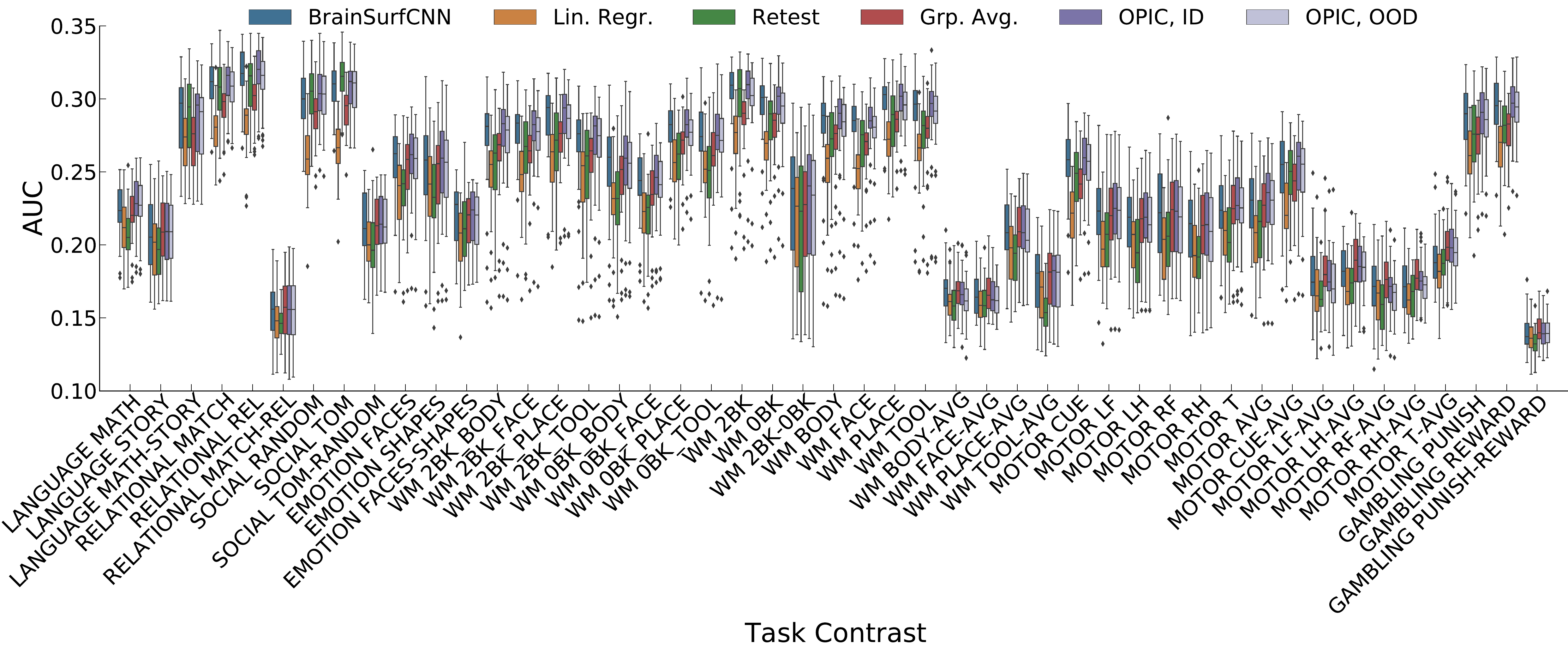}
\caption{Average AUC for all tasks}
\end{figure}

\end{document}